%% file: Template.tex
\documentclass{article}
\usepackage{spconf,amsmath,graphicx}

\usepackage{enumitem}
\usepackage{svg}
\usepackage{hyperref}
\usepackage{amsmath}
\usepackage{amsthm}
\theoremstyle{plain}

\newtheorem*{definition*}{Definition}

\usepackage{color, xcolor}
\usepackage{balance}
\usepackage{subfigure}
\usepackage{array}
\usepackage{graphicx}
\usepackage{booktabs}

\title{TAROT: A Hierarchical Framework with Multitask Co-Pretraining on
Semi-Structured Data towards Effective Person-Job Fit}
\name{
    \begin{tabular}{c}
        Yihan Cao$^{1\star\triangle}$\thanks{$^{\star}$These authors contributed equally to this work.}\thanks{$^{\triangle}$Work performed during the internship at LinkedIn.}
        \qquad Xu Chen$^{2\star\dag}\thanks{$^{\dag}$Corresponding author.}$\qquad Lun Du$^{2}$ 
        \qquad Hao Chen$^{3}$ \qquad Qiang Fu$^{2}$ 
        \qquad Shi Han$^{2}$\\
        Yushu Du$^{1}$\qquad Yanbin Kang$^{1}$
        \qquad Guangming Lu$^{1}$\qquad Zi Li$^{1}$
    \end{tabular}
}

\address{$^{1}$LinkedIn \qquad $^{2}$Microsoft \qquad $^{3}$ Beijing Normal University, China}

\begin{document}
%
\maketitle
\input{src/0_abstract.tex}
\input{src/1_intro.tex}

\input{src/2_method.tex}

\input{src/3_experiment.tex}
\input{src/4_conclusion.tex}

\vfill\pagebreak

\bibliographystyle{IEEEbib}
\bibliography{strings,refs}

\end{document}

%% file: src/0_abstract.tex
\begin{abstract}
Person-job fit is an essential part of online recruitment platforms in serving various downstream applications like Job Search and Candidate Recommendation. Recently, pretrained large language models have further enhanced the effectiveness by leveraging richer textual information in user profiles and job descriptions apart from user behavior features and job metadata. However, the general domain-oriented design struggles to capture the unique structural information within user profiles and job descriptions, leading to a loss of latent semantic correlations. We propose TAROT, a hierarchical multitask co-pretraining framework, to better utilize structural and semantic information for informative text embeddings. TAROT targets semi-structured text in profiles and jobs, and it is co-pretained with multi-grained pretraining tasks to constrain the acquired semantic information at each level. Experiments on a real-world LinkedIn dataset show significant performance improvements, proving its effectiveness in person-job fit tasks.

\end{abstract}

%% file: src/1_intro.tex
\section{Introduction}
\label{sec:intro}

\begin{figure*}[!ht]
    \centering
    \includegraphics[width=0.8\textwidth ]{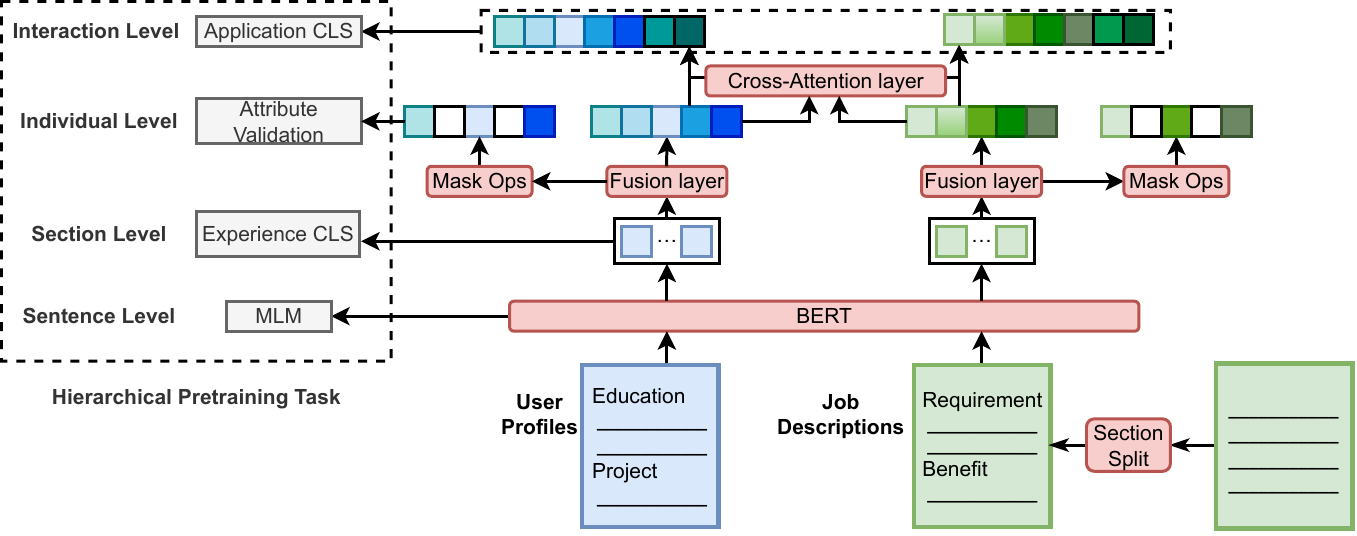}
    \caption{
    The framework of TAROT. Gray boxes refer to different pretraining tasks at each level. }
    \label{structure}
\end{figure*}

Reducing unemployment is a permanent theme for labor and government, especially during epidemic \cite{unemploy}. Enhancing recruitment efficiency, such as Person-Job fit accuracy, can significantly lower unemployment rates, reduce expensive recruitment costs and eliminate job seekers' wasted efforts \cite{humanCaptialReport}.

Traditional efforts on Person-Job fit intent to utilize features from user behaviors or job metadata, like collaborative filtering for job recommendation \cite{zhang2014, lu2013}. The rapid development of online recruitment platforms like LinkedIn and Indeed has facilitated the use of deep learning to obtain text embeddings from large-scale user profiles and job descriptions \cite{qin2018, bian2019}. Recently, large language models (LLMs, e.g., BERT \cite{devlin2019} and GPT-3 \cite{brown2020language}) have proven to be effective in natural language processing and understanding, which become a new choice for learning text representations \cite{bian2020,liao2023concept,Liao2023TexttoImageGF}. However, text organized in structures and domain-specific semantics in Person-Job fit may lead to failures of general domain text-oriented LLMs.


The practical scenario raises new challenges in pretraining LLMs for semi-structured data from specific domains. Firstly, pretrained LLMs are often oriented to general domain corpora, being of low relevance or even contradictory with domain-specific corpora (e.g., abbreviations), leading to embedding collapse of LLMs without domain-specific pretraining \cite{reimers2019sentence,li2020sentence}. Secondly, unlike plain text, texts in user profiles and job descriptions are primarily organized in a domain-specific hierarchical structures since people tend to format them for better illustration of their purposes. Such domain-specific information can undoubtedly promote models in understanding semantics \cite{du2021tabularnet,du2022understanding,chen2020tssrgcn,Chen2021TrafficStreamAS,chen2022graphad}, while it is ignored in current approaches, leaving an improving opportunity.

To tackle these challenges, we propose a hierarchically designed multitask framework \textbf{TAROT} to co-pretrain large language models by incorporating structure information. As Figure \ref{structure} shows, some unstructured job descriptions are segmented by LinkedIn services according to pre-defined sections, constituting recruitment data together with naturally structural user profiles and the rest structured jobs. To match the structure, we elaborate TAROT with corresponding hierarchical structures: sentence $\rightarrow$ section $\rightarrow$ individual $\rightarrow$ interaction level from bottom to top. The semantics of sentences are extracted by BERT, and upper-level embeddings are derived from the attention fusion layer on the lower level. Interaction between job and user embeddings is enhanced via a cross-attention layer so that they can be adaptively adjusted based on needs from the other side. In addition, hierarchical pretraining tasks are designed to encourage integration of structural information and to constrain the model on learning domain-relevant semantics at each level. We use the output embeddings in Person-Job Fit downstream tasks for evaluations. Experimental results on two tasks demonstrate the superiority of TAROT over other baseline methods. The main contributions can be summarized as follows:
\begin{itemize}[leftmargin=*]
\setlength{\itemsep}{0pt}
    \item We propose a hierarchically structured framework for representation learning of person-job fit domain-specific text data as a complement to traditional features. 
    \item We propose multi-grained pretraining tasks specifically for the person-job fit area.
    \item Extensive experiments are conducted to verify the effectiveness of our design and the benefits to downstream tasks.
\end{itemize}

%% file: src/2_method.tex
\section{TAROT}
\label{sec:format}
\subsection{Preliminaries}
We denote the set of users as $U=\{u\}$ and jobs as $J=\{j\}$. 
Job descriptions $j$ are divided into sections $S^j$: Responsibilities, Qualifications, Requirements, Job Title, Functions, Skills, Benefits and Company; and profile sections $S^u$ include Summary, Headline, Education, Position and Skills. Sections consist of sentences like $S^{j}=[s^{j}_{1},\cdots,s^{j}_{k}]$ where $k$ is the number of sentences in $S^j$. The objective of Person-job fit is to predict the matching degree between $u$ and $j$ based on the output embedding of a learned language model.

\subsection{Hierarchical Structured Language Model}
\subsubsection{Language Model}
As a pretraining language model, BERT \cite{devlin2019} has demonstrated promising capabilities in natural language processing tasks in recent years. To empower BERT, TAROT continue-pretrains BERT on large-scale corpora from user profiles and job descriptions on LinkedIn. Sentences are fed into TAROT's language model section-by-section to obtain embeddings.
\subsubsection{Attention Fusion Layer}
As a hierarchically structured model, it is crucial for TAROT to aggregate current level information for the upper level. Section level representations require measuring the importance of different sentences, while individual level embeddings demand distinguishments between sections.
Therefore, we adopt the attention-based fusion method \cite{dai2020} to adaptively learn the difference for embeddings at these two levels. Take profile representation learning at the individual level for example. The embedding sequence of sections is denoted as $E^u = [E^u_*]$. Mathematically, the attention-based fusion embedding is generated as:
\begin{equation}
\begin{split}
    \Tilde{E}^{u}&=\textbf{Pooling}(E^u_*),\\
    E_u'&= \Tilde{E}^{u}+ \textbf{Attention}(Q=\Tilde{E}^{u}, K=E^{u}, V=E^{u}).
\end{split}
\end{equation}

The pooled embedding $\Tilde{E}^{u}$ guides the attention fusion layer to acquire appropriate weights for each section and generate global context-aware individual representations $E_u'$ for user $u$. 

\subsubsection{Cross-Attention Layer}
Empirical evidence suggests that different users will be attracted by different sections of the same job description, and this is similar in the recruiter-profile relationship. 
It inspires us that profile or job description representation learning should not be isolated. Hence, we design the cross-attention layer where the job-oriented attention takes job embeddings $E'_j$ as queries on the profile embeddings $E'_u$ and obtain $A_j$. Similarly, we have $A_u$ from user-oriented attention, and concatenate $A_j$ and $A_u$ as the output. It is then combined with $E'_j$ or $E'_u$ as the final embedding for jobs/user.

\subsection{Hierarchical Pretraining Tasks}
\subsubsection{Sentence-level: Masked Language Model}
Although job descriptions and user profiles are semi-structured data, the sequence of sentences still remains a critical role. Therefore, the classical Masked Language Model (MLM) \cite{devlin2019} is adopted to allow TAROT to emphasize more on the recruitment-related corpora.

\subsubsection{Section-level: Experience Classification}
Given a section $S^{u}$ from user profile $u$, the Experience Classification task is defined as a multi-class classification that utilizes the context under $S^{u}$ to predict its section name from $\{Summary, Headline, Education, Position, Skills\}$. Technically, sentences of $S^{u}$ will be fed into BERT to get the representations, and then the output of the entire section is taken as the input of a Multi-Layer Perceptron (MLP) to predict the label of section name $\Tilde{y}_{S^u}$. Experience Classification is to minimize the cross-entropy loss between $\Tilde{y}_{S^u}$ and the real section name label $y_{S^u}$ of $n_{Exp}$ samples. For convenience, the objective is formulated as Eq. \ref{eq1}, e.g., Experience Classification objective is $L_{Exp}(n_{Exp}, \Tilde{y}_{S^u}, y_{S^u})$:
\begin{equation}\label{eq1}
L_{task}(n_{task}, \Tilde{y}, y)=-\frac{1}{n_{task}}\sum y\cdot\log \Tilde{y} + (1-y)\cdot\log(1-\Tilde{y}).
\end{equation}

\subsubsection{Individual-level: Attribute Validation}
To focus on some co-occurring attributes for embeddings on both sides, we carefully design the attribute validation task at the individual level so that key attributes are better incorporated. The attribute validation task leverages embeddings from previous layers to predict attributes of users and jobs. Here \textit{Skill} is chosen as the key individual information. The reason is that under the person-job fit scenario, a match between a job and a user profile highly depends on the skills commanded by the user and required by the job. The label is the unified ``skill\_ids'' that are extracted by LinkedIn service from the skills section. 
Representations for attribute validation are obtained from the individual-level attention fusion layer in two steps: skill section will be removed from $\Tilde{E}^u$ during pooling (denoted as $\Tilde{E}^u_{msk}$), and it is also masked when conducting self-attention:
\begin{equation}
E'_{u,msk}\!\!=\! \Tilde{E}^u_{msk} \!+\! \textbf{Attention}(Q\!=\!\Tilde{E}^u_{msk}, K\!=\!E^u,V\!=\!E^u).
\end{equation}
Predicted label $\Tilde{y}_{Att}$ is generated through a single layer MLP on $E'_{u,msk}$. 
Multi-label one-versus-all loss is used to transform it into a series of binary classification problems, and the Attribute Validation objective can be denoted as $L(n_{Att}, \Tilde{y}_{Att}, y_{Att})$.
A similar objective can be defined for job representations.

\subsubsection{Interaction-level: Application Classification}
Application Classification task is designed to predict whether a user will apply for a job in order to strengthen interactions between user profile embedding and job description embedding. 
It is worth mentioning that negative samples are randomly generated and accounts for 3/4 of the dataset. The reason is that the collected application data highly rely on job recommendations, and users only react to recommended jobs that are already considered to be suitable ones in the system. 
Practically, when recommended a job, the user can choose to skip/dismiss/save/apply for this job. The ``skip'' action is excluded while ``apply'' and ``save'' are considered positive and ``dismiss'' is negative.
The final output of TAROT will be fed into a single layer MLP to generate the predicted label $\Tilde{y}_{App}$ and trained with corresponding object $L(n_{App}, \Tilde{y}_{App}, y_{App})$.

\subsection{Pretraining and Downstream Evaluation}
Text from job descriptions and user profiles are co-pretrained with hierarchical tasks, and the overall objective can be formulated as: 
\begin{equation}
    L=\sum_{*} \lambda_{*}L_{*},\text{where }*\in\{MLM,Exp,Att,App\},
\end{equation}
where $\lambda_*$ is the hyper-parameter for the corresponding pretraining task. We select two downstream tasks including job recommendations for users and user recommendations for jobs (recruiters).

%% file: src/3_experiment.tex
\section{Experiments}
\label{sec:pagestyle}

\begin{table*}[]
\centering
   \resizebox{0.99\textwidth}{!}{%
  \begin{tabular}{l| ccccc | ccccc}
    \toprule
    \textbf{Task} & \multicolumn{5}{c|}{\textbf{Job Recommendation}} & \multicolumn{5}{c}{\textbf{Candidate Recommendation}}\\\midrule
    \textbf{Models} & \textbf{AUC} & \textbf{Recall@3} & \textbf{Precision@3} & \textbf{NDCG@3} & \textbf{MRR} & \textbf{AUC} & \textbf{Recall@3} & \textbf{Precision@3} & \textbf{NDCG@3} & \textbf{MRR}\\
    \midrule
    PJFNN & - & - & - & - & - & - & - & - & - & -  \\ 
    PJFNN+Bert & +2.538\% & +2.280\% & +0.899\% & +2.027\% & +0.582\% & +0.334\% & +1.464\% & +1.569\% & +1.768\% & +0.735\%  \\
    \textbf{PJFNN+TAROT} & \textbf{+4.477\%} & \textbf{+3.941\%} & \textbf{+4.780\%} & \textbf{+3.896\%} & \textbf{+3.831\%} & \textbf{+6.977\%} & \textbf{+8.176\%} & \textbf{+9.765\%} & \textbf{+13.494\%} & \textbf{+9.007\%}  \\ \midrule
    BPJFNN & +4.658\% &  -0.732\% & +2.130\% &  -1.211\% &  -0.436\% & +2.765\% & +2.990\% &  -0.959\% & +2.475\% & +0.827\%  \\
    BPJFNN+Bert & +5.057\% & +0.929\% & +3.739\% &  -0.211\% & +1.334\% & +3.229\% & +4.637\% & +1.831\% & +4.537\% & +2.987\%  \\
    \textbf{BPJFNN+TAROT} & \textbf{+6.163\%} & \textbf{+4.110\%} & \textbf{+6.152\%} & \textbf{+1.003\%} & \textbf{+3.177\%} & \textbf{+7.385\%} & \textbf{+10.189\%} & \textbf{+8.718\%} & \textbf{+14.555\%} & \textbf{+7.721\%}  \\ \midrule
    APJFNN & +9.679\% & +3.941\% & +6.294\% & +0.684\% & +3.637\% & +6.198\% & +7.444\% & +12.119\% & +11.314\% & +6.985\%  \\
    APJFNN+Bert & +10.694\% & +4.899\% & +7.856\% & +2.237\% & +4.680\% & +6.866\% & +9.457\% & +15.606\% & +12.080\% & +7.675\%  \\
    \textbf{APJFNN+TAROT} & \textbf{+11.891\%} & \textbf{+6.278\%} & \textbf{+10.554\%} & \textbf{+4.396\%} & \textbf{+5.941\%} & \textbf{+8.295\%} & \textbf{+14.216\%} & \textbf{+18.309\%} & \textbf{+17.030\%} & \textbf{+10.386\%} \\
    \bottomrule
  \end{tabular}
  }
  \caption{Performance comparison on two downstream tasks. Results are relative improvements compared to PJFNN.}
  \label{performance1}
  \vspace{-0.3cm}
\end{table*}

\subsection{Experiment Settings}
\begin{table}[h]
  \resizebox{0.45\textwidth}{!}{%
  \begin{tabular}{l |ccccc}
    \toprule
    \textbf{Task} & \multicolumn{5}{c }{\textbf{Job Recommendation}}\\\midrule
    \textbf{Models} & \textbf{AUC} & \textbf{HR@1} & \textbf{NDCG@5} & \textbf{NDCG@25} & \textbf{MRR}\\
    \midrule
    w/o MLM & -4.9\% & -11.0\%& -8.3\%& -10.0\%& -8.7\%\\
    w/o EXP & -5.1\% & -6.8 \%& -4.5\%& -8.0 \%&-8.7\%\\
    w/o ATT & -4.6\% & -2.3 \%& -4.9\%& -2.5 \%&+0.4\%\\
    w/o APP & -9.5\% & -13.7\%&-11.3\%& -9.3 \%&-10.5\%\\
    \bottomrule
  \end{tabular}}
  \caption{Multitask ablation study with TAROT as baseline.}
  \label{ablation}
  \vspace{-0.5cm}
\end{table}

\begin{table}[h]
  \resizebox{0.45\textwidth}{!}{%
  \begin{tabular}{l |ccc|ccc}
    \toprule
    \textbf{Task} & \multicolumn{3}{c|}{\textbf{Job Recommendation}} & \multicolumn{3}{c}{\textbf{Candidate Recommendation}} \\ \midrule
    \textbf{Models} & \textbf{AUC} & \textbf{NDCG@5} & \textbf{MRR} &\textbf{AUC} & \textbf{NDCG@5} & \textbf{MRR}\\
    \midrule
    OF & - & - & - & - & - & - \\
    OF+BERT  & +0.3\% & -0.7 \% & -1.9\% & +1.8\% & +0.6\% & +0.6\% \\
    OF+TAROT & +6.0\% & +12.9\% & +6.0\% & +5.5\% & +4.3\% & +4.5\% \\
    \bottomrule
  \end{tabular}}
  \caption{Improvement to online service features. ``OF'' refers to online features used in LinkedIn system.
  }
  \label{RW_plugin}
  \vspace{-0.5cm}
\end{table}

\noindent\textbf{Dataset}
The data for pre-training is collected from user activity records of LinkedIn with anonymized user profiles and job descriptions for security. We filter out incomplete user profiles, and the training data contains over 800k job application records, including 193k users and 331k jobs. There are two downstream tasks. 

For \textbf{job recommendation} task, the data is composed of job and user profile pairs. 
``Skip'' and ``Dismiss'' actions are labeled negative, and ``Save" and ``Apply" are positive. The dataset contains 31k users and 54k jobs, with 150k samples.
The \textbf{candidate recommendation} task recommends users to recruiters according to their posting jobs. When a user is recommended, the recruiter can contact or skip the candidate. The dataset contains 133k users and 19k jobs, with 150k samples.
To prevent data leakage, we select a different user group than the training data in the data of these two tasks. 

\noindent\textbf{Compared Method}
We compare TAROT with variants of PJFNNs to examine the benefits of extracting semi-structured text. Hence, the compared methods are divided into three types: (1) \textbf{PJFNNs} \cite{qin2018, bian2019} are a series of models that are proven to be effective for person-job fit; 
(2) \textbf{PJFNNs + BERT} 
 that uses BERT as a plugin for semantic embeddings. (3) \textbf{PJFNNs + TAROT} utilizes TAROT embeddings that is similar to PJFNNs+BERT. 

\noindent\textbf{Implementation Details}
We choose the small BERT model \cite{smallbert} with 512 hidden neurons.
Adam is utilized as the optimizer and the learning rate is $10^{-4}$. 
We use grid search strategy to find best hyper-parameters.
AUC, top-3 recall rate (Recall@3), precision rate (Precision@3), Normalized Discounted Cumulative Gain (NDCG@3) and Mean Reciprocal Rank (MRR) are used to evaluate model performance. We take PJFNN as the foundation and other results are relative improvements against it.
\subsection{Downstream Performance Comparison}

The results are organized in Table \ref{performance1}. Compared to PJFNNs, additional BERT embeddings can provide improvements on most metrics, showing the superiority of incorporating semantics information from pretrain LLMs. We can also discover that TAROT significantly enhance the recommendation performance on both tasks, demonstrating that embeddings from the hierarchically co-pretraining framework are more expressive and informative than the pretrain-based semantic plugin BERT. 
Besides, TAROT not only achieves remarkable performance on the job recommendation task that is highly correlated with the pretraining Application Classification task, but also obtains impressive gains on the candidate recommendation task. Note that ``headhunter'' is not explicitly included in our framework and headhunters will post numerous job so it is also not in the individual-level. However, the candidate recommendation task is headhunter-oriented, which implies that even though there is no corresponding pretraining task, TAROT embeddings can still be beneficial to the generalization of models to unseen Person-Job fit downstream tasks.

\subsection{Ablation Study}
To evaluate the design of multitask training in our framework, we conduct ablation studies by removing different pretraining tasks, and observe the performance on the downstream job recommendation task as there are corresponding entire pretrain task sets. Here we add the top-1 hit rate (HR@1) metric and the results are shown in Table \ref{ablation}. ``w/o App'', ``w/o Att'', ``w/o Exp'' and ``w/o MLM'' refer to pretraining without the corresponding task.
From the results, we can see that Experience Classification and Attribute Validation are essential to our downstream tasks as removing them will degrade the performance. The worst result in w/o App indicates that Application Classification plays the most critical role because it further empowers information interactions between job descriptions and user profiles. In summary, all the results prove the effectiveness of our multitask co-pretraining framework.
\subsection{Improvement to Online Service Features}
We also combine TAROT embeddings with features used in LinkedIn online service. From Table. \ref{RW_plugin} we can find that TAROT embeddings can provide additional gains to currently-used features and are more effective than BERT, which implies its value in practice. Notably, individual-level embeddings can be stored to speed up the inference in online products.

%% file: src/4_conclusion.tex
\section{Conclusion}

In this paper, we propose TAROT to provide expressive embeddings for person-job fit applications. To fully leverage the text and interaction information from job descriptions and 
user profiles, we design a hierarchical multitask co-pretraining framework for a better understanding of the semantic information and correlations of them. To evaluate the effectiveness, we conduct comprehensive experiments on the real data of LinkedIn with several baselines. The experimental results show that our framework can significantly improve downstream task performance and promote the online service feature in LinkedIn.